# Fiber Tract Shape Measures Inform Prediction of Non-Imaging Phenotypes

Wan Liu[1,2], Yuqian Chen[2,3], Chuyang Ye[1], Nikos Makris[2], Yogesh Rathi[2], Weidong Cai[3], Fan Zhang[2], Lauren J. O'Donnell[2]

[1] School of Integrated Circuits and Electronics, Beijing Institute of Technology, Beijing, China
[2] Harvard Medical School, Boston MA, USA
[3] The University of Sydney, NSW, Australia
odonnell@bwh.harvard.edu

**Abstract.** Neuroimaging measures of the brain's white matter connections can enable the prediction of non-imaging phenotypes, such as demographic and cognitive measures. Existing works have investigated traditional microstructure and connectivity measures from diffusion MRI tractography, without considering the shape of the connections reconstructed by tractography. In this paper, we investigate the potential of fiber tract shape features for predicting non-imaging phenotypes, both individually and in combination with traditional features. We focus on three basic shape features: length, diameter, and elongation. Two different prediction methods are used, including a traditional regression method and a deep-learning-based prediction method. Experiments use an efficient two-stage fusion strategy for prediction using microstructure, connectivity, and shape measures. To reduce predictive bias due to brain size, normalized shape features are also investigated. Experimental results on the Human Connectome Project (HCP) young adult dataset (n=1065) demonstrate that individual shape features are predictive of non-imaging phenotypes. When combined with microstructure and connectivity features, shape features significantly improve performance for predicting the cognitive score TPVT (NIH Toolbox picture vocabulary test). Overall, this study demonstrates that the shape of fiber tracts contains useful information for the description and study of the living human brain using machine learning.

## 1    Introduction

Recent research demonstrates the potential of better understanding the brain in health and disease by investigating how neuroimaging data (such as MRI) relates to non-imaging phenotypes (such as sex, age, and behavioral and clinical variables) [1]. One popular strategy is the development of machine learning methods to relate these high-dimensional sources of information (e.g., [2,3]). Diffusion MRI (dMRI) tractography, the only method that can reconstruct the brain's white matter connections or fiber tracts in vivo, has contributed to predicting non-imaging phenotypes (e.g., [2,4–7]). However, these prediction methods focus on a subset of the information available from dMRI tractography. Most dMRI tractography research relies on tissue microstructure measures such as fractional anisotropy, or connectivity "strength" measures between pairs of gray matter regions [8].

A limited body of work focuses on another inherent property of the brain's white matter connections: their shape. For example, one recent study demonstrated that shape measures such as length and diameter describe the variability of association tracts in a population [9]. The volume of fiber tracts has also been studied, with characteristic changes across the lifespan [10]. Recently, information about the cross-sectional area of brain connections was shown to enhance correspondence between structural and functional connectivity [11]. Other shape measures, such as curvature and torsion [12] or fiber dispersion [13], have been proposed to describe local shapes along fiber tracts. These understudied shape features hold promise for investigating the relationship between the brain's connectional anatomy and non-imaging phenotypes. In this initial work, we investigate the predictive potential of several basic shape features in combination with traditional microstructure and connectivity features. We focus on machine learning testbeds for predicting age, sex, and example cognitive measures from the NIH toolbox [14].

The main contribution of this paper is to demonstrate the potential of fiber tract shape features for predicting non-imaging phenotypes, both individually and in combination with traditional features. We also investigate the predictive potential of normalized shape measures to reduce the effect of brain size, a known confound for brain-based prediction [15]. Two different prediction methods are used in the experiments for comparison, including a traditional regression method and a deep-learning-based prediction method. Overall, we aim to show that shape features provide a new way to study the brain's white matter connections using machine learning.

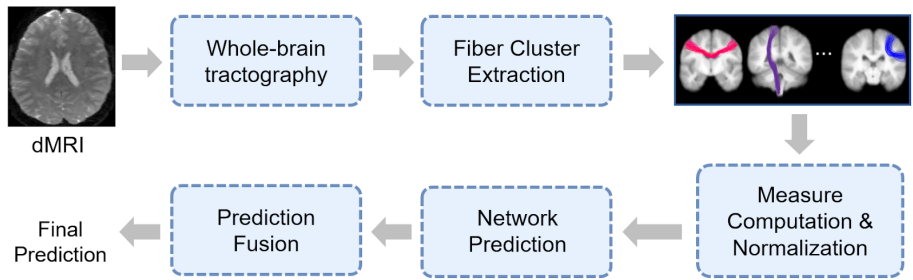

**Fig. 1.** Overview of the proposed method.

## 2    Methodology

An overview of the proposed method is shown in Fig. 1, including computation of whole-brain tractography, white matter parcellation, extraction of fiber tract measures (i.e., microstructure, connectivity, and shape), and prediction of non-imaging phenotypes.

### 2.1    Data Processing and Cognitive Measures

We used dMRI and demographic/behavioral data from 1065 subjects (490 female, 22-37 years old) in the Human Connectome Project (HCP) young adult dataset [16]. The following processing steps were performed for tractography and parcellation of the white matter. We computed whole-brain tractography with the two-tensor Unscented Kalman Filter (UKF) method [17], which fits a mixture model of two tensors to the dMRI during fiber tracking. Then, the whitematteranalysis (WMA) tool was utilized to extract fiber clusters based on an existing white matter (WM) atlas [18]. For each subject, we obtained 716 clusters in each brain hemisphere and 84 commissural clusters, resulting in a total of 1516 fiber clusters per subject. Multiple measures from each fiber cluster (see Section 2.2) were then computed to serve as input for prediction tasks.

We considered five different prediction tasks in our work: one classification task for sex prediction, and four regression tasks to predict age and cognitive scores. The NIH Toolbox cognitive scores [19] include TPVT (Toolbox Picture Vocabulary Test to assess vocabulary comprehension ability), TORRT (Toolbox Oral Reading Recognition Test to assess reading decoding skill and crystallized ability), and TFAT (Toolbox Flanker Inhibitory Control and Attention Test to assess visuospatial attention). TPVT scores range from 90.7 to 153.1 (117.1±9.6). TORRT scores range from 84.2 to 150.7 (117.0±10.6). TFAT scores range from 85.7 to 142.1 (111.8±10.1).

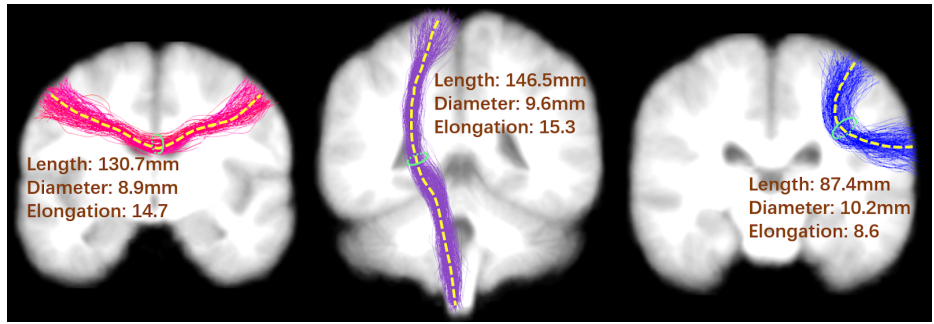

**Fig. 2.** Example fiber clusters illustrating the shape measures computed in this study.

### 2.2    Measure Estimation and Normalization

We extracted measures of fiber cluster microstructure, connectivity, and shape. Microstructure measures included fractional anisotropy (FA) and mean diffusivity (MD), computed by averaging the FA/MD values of all points in a fiber cluster [20]. Connectivity measures included the number of streamlines (NoS). Shape measures (see Fig. 2) were computed as follows.

**Length:** This is an important characteristic of fiber pathways [9,21] that relates to conduction speed [22]. We computed the mean streamline length of each cluster in mm.

**Diameter:** Several measures of diameter have been proposed for fiber tracts including the mean diameter of all cross-sections along the fiber tract [9] and the estimated total axonal cross-section [11]. In this work, we measured the diameter of the cross-section of the central part of the fiber cluster. We estimated the diameter using the covariance matrix of the midpoints of the streamlines within a fiber cluster. The cluster cross-sectional diameter was estimated using the maximum eigenvalue $e_{max}$ of this covariance matrix [23]:

$$diameter = 2\sqrt{e_{max}}$$ (1)

**Elongation:** This measure is defined as the ratio of the mean length and diameter of a fiber cluster [9]. It is a dimensionless measure that was shown to have high test-retest reliability for describing white matter association connections [9].

**Normalization for brain size:** Brain size is known to affect streamline length [24] and connectivity measures [25], and to strongly affect machine learning prediction of sex [15]. To investigate the effect of normalization on prediction tasks, we followed standard practice to reduce the effect of brain size by dividing by a subject-specific reference value [26]. We used the subject-specific mean across clusters as a reference value for the normalization of each connectivity and shape measure.

### 2.3    Network Structure

We implemented a simple yet effective pipeline using a 1D-CNN method that has been shown to have good performance for prediction tasks using dMRI tractography features [27,28]. It consists of a feature extractor and a classifier. The feature extractor contains three convolutional blocks, and in each block there is a 1D convolutional layer (kernel size=5, kernel number=64, stride=1), a batch normalization layer, and a ReLU activation layer. The classifier is composed of two fully connected layers, and the output dimensions are 512 and 128, respectively [27]. The input of the 1D-CNN is a vector of length 1516 containing a measure (Section 2.2) for each fiber cluster.

For comparison on regression tasks, we also included results from ElasticNet (E-Net), which is a popular regression method for the prediction of non-imaging phenotypes [2,7,29]. ElasticNet is a linear regression approach in machine learning that regularizes the regression models with both the lasso and ridge techniques [30].

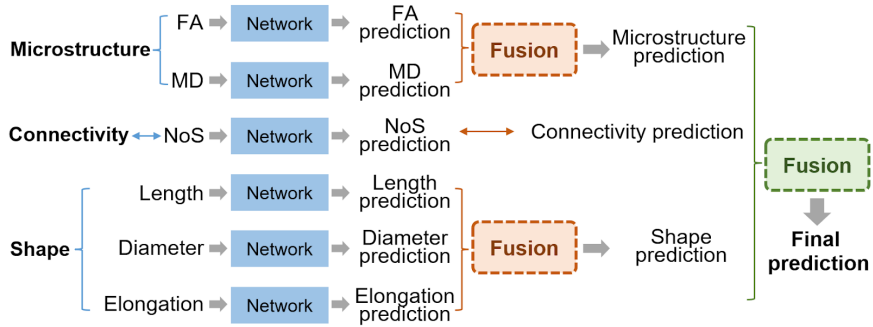

**Fig. 3.** Two-stage prediction fusion of microstructure, connectivity, and shape measures.

### 2.4     Two-stage Fusion Strategy for Microstructure, Connectivity, and Shape

We designed a two-stage fusion strategy to fuse results within and across the microstructure, connectivity, and shape measures (see Fig. 3). This fusion strategy lets us assess the performance of the shape measures both independently and in combination with other measures, for various prediction tasks. We first independently trained an initial network for each measure from Section 2.2. We then performed a within-category measure fusion to produce separate microstructure-specific, connectivity-specific, and shape-specific networks. Finally, in the cross-category fusion stage, we ensembled the results across the networks of different categories. We used appropriate fusion methods (voting or averaging) for each classification or regression task.

### 2.5     Implementation Details

All experiments in this work were performed on an NVIDIA GeForce GTX 1080 Ti GPU based on Pytorch. Like [27], 1D-CNN was trained with the Stochastic Gradient Descent (SGD) optimization algorithm for 300 epochs, and the learning rate and batch size were set to 0.1 and 8, respectively. The alpha value of ElasticNet regression is set to 0.005 after tuning with each candidate in {1,0.5,0.1,0.05,0.01,0.005,0.001} [31]. Cross entropy and mean squared error are used as the loss functions for the classification and regression tasks, respectively. For any missing clusters due to intersubject variability, all of the measures were set to zero [27]. The diameter and elongation were set to zero if only one streamline is included in the cluster. We utilized max-min normalization for each measure within a subject during network training and testing, as commonly used in deep learning methods [32].

## 3     Experiments and Results

For both prediction methods (1D-CNN and E-Net), we performed 5-fold cross-validation for experimental evaluation, in which each validation trial used 4/5 of the 1065 HCP subjects for training and the remaining 1/5 for testing. Following the literature, performance on prediction of sex, age, and cognitive function was reported with metrics of accuracy (Acc), mean absolute error (MAE), and Pearson's correlation coefficients (r), respectively [1,33]. For statistical comparison across methods, a one-way repeated measures (ANOVA) test was performed, followed by post-hoc pairwise comparisons using paired t-tests.

### 3.1     Evaluation of Individual Measures for Prediction

The prediction performance of individual measures is shown in Table 1. Overall, it can be observed that fiber cluster shape measures are informative for prediction. For example, all the shape measures achieve around 90% accuracy for sex prediction, and compared with the microstructure measures, the diameter, elongation, and normalized diameter shape measures have comparable accuracy for predicting TPVT and higher accuracy for predicting TFAT. However, for most prediction tasks the performance of shape measures is lower than connectivity measures. As expected, the performance of the normalized shape measures is often lower than the original shape measures, due to the reduction of bias due to brain size. This reduction is small, indicating that both sets of measures may be useful for the prediction tasks.

**Table 1.** Prediction performance of individual measures. '-N' indicates measure normalization. The best performance for each task is shown in bold.

| Category | Measure | Sex (Acc%) | Age (MAE) | TPVT (r) | TORRT (r) | TFAT (r) |
|---|---|---|---|---|---|---|
| Microstructure | FA | **94.33±0.79** | 2.82±0.11 | 0.33±0.04 | 0.29±0.05 | 0.20±0.03 |
| | MD | 92.69±2.91 | **2.70±0.14** | 0.32±0.10 | 0.31±0.05 | 0.18±0.04 |
| Connectivity | NoS | 94.04±2.60 | 2.77±0.11 | **0.37±0.05** | **0.34±0.06** | 0.20±0.03 |
| | NoS-N | 94.23±2.72 | 2.79±0.11 | **0.37±0.07** | **0.34±0.06** | 0.22±0.04 |
| Shape | Length | 89.90±2.04 | 2.96±0.09 | 0.22±0.03 | 0.14±0.06 | 0.14±0.04 |
| | Diameter | 90.77±3.42 | 3.00±0.12 | 0.33±0.07 | 0.25±0.09 | 0.22±0.04 |
| | Elongation | 90.19±3.29 | 2.99±0.11 | 0.28±0.05 | 0.24±0.05 | 0.21±0.05 |
| | Length-N | 89.14±3.01 | 2.99±0.09 | 0.21±0.06 | 0.16±0.03 | 0.16±0.03 |
| | Diameter-N | 90.48±3.21 | 2.99±0.08 | 0.32±0.08 | 0.24±0.09 | **0.24±0.03** |
| | Elongation-N | 90.58±3.44 | 3.09±0.10 | 0.26±0.04 | 0.23±0.04 | 0.15±0.07 |

**Table 2.** Performance of within-category fused networks. '-N' indicates measure normalization. The best performance for each task is shown in bold.

| Category | Sex (Acc%) | Age (MAE) | TPVT (r) | TORRT (r) | TFAT (r) |
|---|---|---|---|---|---|
| Microstructure | **94.69±1.27** | **2.64±0.12** | 0.35±0.03 | 0.31±0.04 | 0.20±0.05 |
| Connectivity | 94.04±2.60 | 2.77±0.11 | **0.37±0.05** | **0.34±0.06** | 0.20±0.03 |
| Connectivity-N | 94.23±2.72 | 2.79±0.11 | **0.37±0.07** | **0.34±0.06** | 0.22±0.04 |
| Shape | 91.83±2.40 | 2.92±0.09 | 0.32±0.05 | 0.23±0.05 | **0.23±0.05** |
| Shape-N | 92.40±2.36 | 2.95±0.09 | 0.34±0.07 | 0.26±0.07 | 0.22±0.05 |

### 3.2    Evaluation of Microstructure, Connectivity, and Shape Networks

Compared with the result of individual measures in Table 1, within-category fusion (to produce Microstructure, Connectivity, and Shape Networks) consistently improves prediction performance (Table 2). Again, the fused shape measures are predictive for these tasks, and shape measures achieve the best performance on TFAT, with somewhat lower performance on other prediction tasks than traditional measures. Interestingly, the fused Shape-N measures slightly outperform the Shape measures on several tasks, again supporting that both sets of shape measures have potential utility for the study of the brain.

### 3.3    Evaluation of Overall Fused Networks

Compared with the within-category fused networks in Table 2, it can be seen that the second stage cross-category fusion improves performance (Table 3). Shape features significantly improve performance on the task of TPVT prediction. TPVT prediction performance is significantly different across fused networks (ANOVA $p$=0.032), and the fused {Microstructure, Connectivity, Shape} 1D-CNN significantly outperforms the {Microstructure, Connectivity} 1D-CNN (paired t-test, $p$=0.039). This indicates that shape features contain information related to language performance. Furthermore, this information is complementary to the information contained in traditional microstructure and connectivity features. The other fused 1D-CNN networks in Table 3 all have similar performance (no significant difference found with repeated measures ANOVA) on tasks of sex, age, TORRT, and TFAT prediction.

We also utilize ElasticNet (E-Net) for comparison (Table 3). As expected, E-Net performance is generally lower than 1D-CNN. TPVT results are consistent: performance is significantly different across networks for TPVT prediction (ANOVA $p$=0.024), and the fused {Microstructure, Connectivity, Shape} E-Net significantly outperforms the {Microstructure, Connectivity} E-Net (paired t-test, $p$=0.033). This indicates that for both networks the TPVT prediction performance can be consistently improved by fusing the shape measures with the traditional microstructure and connectivity measures. Again, the fused E-Nets obtain similar performance for TFAT prediction (no significant difference for ANOVA). On age and TORRT prediction tasks, ANOVA indicates a significant difference in performance across fused networks, which can be attributed to a significant change in prediction performance related to normalization (of Shape-N for age prediction and Connectivity-N for TORRT prediction).

**Table 3.** Performance of cross-category fused networks. '-N' indicates measure normalization. Significantly better performance after the addition of shape measures is shown in bold. (n.s.:$p$>0.05; *:$p$<0.05; **:$p$<0.01;***:$p$<0.001)

| Categories | Method | Sex (Acc%) | Age (MAE) | TPVT (r) | TORRT (r) | TFAT (r) |
|---|---|---|---|---|---|---|
| Microstructure Connectivity | 1D-CNN | 95.48±2.51 | 2.65±0.10 | 0.39±0.04 | 0.35±0.04 | 0.22±0.01 |
| | E-Net | – | 2.78±0.09 | 0.31±0.04 | 0.26±0.03 | 0.12±0.03 |
| Microstructure Connectivity-N | 1D-CNN | 94.71±2.71 | 2.68±0.08 | 0.39±0.04 | 0.35±0.06 | 0.21±0.01 |
| | E-Net | – | 2.82±0.09 | 0.32±0.03 | 0.29±0.04 | 0.13±0.04 |
| Microstructure Connectivity Shape | 1D-CNN | 95.48±2.22 | 2.67±0.09 | **0.42±0.04\*** | 0.35±0.05 | 0.23±0.02 |
| | E-Net | – | 2.81±0.09 | **0.33±0.05\*** | 0.27±0.05 | 0.14±0.04 |
| Microstructure Connectivity-N Shape-N | 1D-CNN | 95.48±2.03 | 2.682±0.12 | 0.407±0.05 | 0.34±0.04 | 0.22±0.04 |
| | E-Net | – | 2.88±0.09 | 0.33±0.04 | 0.29±0.05 | 0.16±0.08 |
| ANOVA | 1D-CNN | n.s. | n.s. | * | n.s. | n.s. |
| | E-Net | – | ** | * | *** | n.s. |

## 4    Discussion and Conclusion

In this paper, we investigated the potential of three basic shape features, in combination with traditional microstructure and connectivity features, for the prediction of several non-imaging phenotypes using dMRI tractography data. We found that the utility of individual shape features for prediction was high and that the addition of shape information could significantly improve the prediction of language performance (TPVT). Future work should investigate additional shape measures that have been proposed in the literature such as curvature [12], fiber dispersion [13], and area of intersection with the cortex [9]. Overall, our results demonstrate that the shape of fiber tracts contains useful information for the description and study of the living human brain using machine learning.

**Acknowledgments** We acknowledge the following NIH grants: P41EB015902, R01MH074794, R01MH125860 and R01MH119222.